\documentclass{article}

\usepackage{arxiv}

\usepackage[utf8]{inputenc} % allow utf-8 input
\usepackage[T1]{fontenc}    % use 8-bit T1 fonts
\usepackage{hyperref}       % hyperlinks
\usepackage{url}            % simple URL typesetting
\usepackage{booktabs}       % professional-quality tables
\usepackage{amsfonts}       % blackboard math symbols
\usepackage{nicefrac}       % compact symbols for 1/2, etc.
\usepackage{microtype}      % microtypography
\usepackage{lipsum}		% Can be removed after putting your text content
\usepackage{dsfont}

\usepackage{amsmath}

\DeclareMathOperator*{\argmin}{arg\,min}
\usepackage{algorithmic}
\usepackage{algorithm}
\usepackage{bbm}
\usepackage{graphicx}

\newtheorem{proposition}{Proposition}
\newtheorem{proof}{Proof}
\newtheorem{lemma}{Lemma}

\title{Tackling Algorithmic Bias in Neural-Network Classifiers using Wasserstein-2 Regularization}

\author{Laurent Risser$^{1,2}$, Alberto Gonzalez Sanz$^{1,2}$, Quentin Vincenot$^{3}$, Jean-Michel Loubes$^{1,2,3}$}

\date{
  $\null^{1}$ Institut de Math\'ematiques de Toulouse (UMR 5219), CNRS, Université de Toulouse, F-31062 Toulouse, France\\
  $\null^{2}$ Artificial and Natural Intelligence Toulouse Institute (ANITI), Toulouse, France\\
  $\null^{3}$ Institut de Recherche Technologique (IRT) Saint Exup\'ery, Toulouse, France\\
}

\begin{document}
\maketitle

\begin{abstract}
The increasingly common use of neural network classifiers in industrial and social applications of image analysis has allowed impressive progress these last years. Such methods are however sensitive to algorithmic bias, \textit{i.e.} to an under- or an over-representation of positive predictions or to higher prediction errors in specific subgroups of images.
We then introduce in this paper a new method to temper the algorithmic bias in Neural-Network based classifiers. Our method is Neural-Network architecture agnostic and  scales well to massive training sets of images.
It indeed only overloads the loss function with a Wasserstein-2 based regularization term
for which we back-propagate the impact of specific output predictions using a new model, based on the G\^ateaux derivatives of the predictions distribution.
This model is algorithmically reasonable and makes it possible to use our regularized loss with standard stochastic gradient-descent strategies.
Its good behavior is  assessed on the reference \textit{Adult census}, \textit{MNIST}, \textit{CelebA} datasets.
\end{abstract}

% keywords can be removed

\section{Introduction}

\subsection{Algorithmic bias in Machine Learning}\label{ssec:AlgorithmicBiasMachineLearning}

Recent neural-network models have become extremely popular for a large variety of  applications in image analysis. They have indeed made it possible to strongly improve the accuracy of classic image analysis strategies for the detection of specific shapes, as originally shown in \cite{LeCunNC1989,LeCunIEEEproc1998}, or other applications of image analysis. A huge effort has then been made to design neural network architectures with desirable properties  \cite{KrizhevskyNIPS2012,He2016DeepRL,BengioPAMI2013} and to efficiently optimize  their parameters \cite{Kingma14,DauphinNIPS2014}.
How powerful they might be, these models may however suffer from algorithmic bias.
This bias impacts the predictions, generally in the detriment of a given subset of observations. For instance, the prediction errors may be higher in a distinct  images subset compared than in the other images. The portion of predictions of a given label may also be particularly low in a given subgroup of images with no valid reason. The effect of this bias in the society has been studied recently in various papers \cite{JohndrowAAS2019,buolamwini2018gender,Besse2020_BiasSurvey}. In this paper, we focus on a new strategy to tackle it when detecting specific image features using neural networks.

A popular illustration of the effects of algorithmic bias on images was shown in \cite{buolamwini2018gender}, where the authors measured that different publicly available commercial face recognition online services achieved their lowest accuracy on females. For instance, Face++ had a 99.3\% accuracy to recognize male politicians and only a 78.7\% for female  politicians, so about 95.9\% of the errors were obtained on females. In the general case, the literature dealing with algorithmic bias in Machine Learning considers that the bias comes from the impact of a non-informative variable denoted \textit{sensitive (or protective) variable}. This variable is the \textit{gender} in the former example. It splits the observations into two subsets for which the Machine Learning predictions have different behaviors, implying that  they are biased with respect to this variable.
These biases are widely discussed in the \textit{Fair learning} community, which studies their effects in the society. Because of the increasingly massive use of machine learning-based applications in the everyday life, justice, or resource allocations, this community has become extremely dynamic, as shown by the strong  emergence of a conference like  \textit{ACM FAT*}\footnote{https://www.fatml.org/}. The Computer Vision community has also drawn a significant attention to such questions recently with for instance the contribution of \cite{Quadrianto_CVPR2019}, the organization of the CVPR 2018 workshop \textit{Computer vision meets fairness}, and the CVPR 2019 workshop \textit{Fairness Accountability Transparency and Ethics in Computer Vision}.

To make obvious that these biases are common in  Machine-Learning based applications of image analysis, we detail hereafter three major causes of algorithmic bias in Machine-Learning:
\textbf{(Cause 1)}
Machine  learning  algorithms are first meant to automatically take accurate and efficient decisions that mimic human expertise, based on reference datasets that are potentially biased with respect to one or several protective variables. An algorithm is then likely to reproduce the errors contained in a biased training database, although there is generally no intention of doing so.  A biased training set, may be due to specific persons, who label the training data with little experience. It may also be due to a reference algorithm which labels automatically the data with a bias on a specific type of data. In the fair-learning community, it is often considered as being due to societal biases in the sampled populations or to systematic biases made by the persons who took the reference decisions.
Note that these errors are considered as systematic in the sensitive sub-group, at least in law.
\textbf{(Cause 2)} The algorithmic bias may also be due to the fact that a classifier overfits or underfits a specific sub-group of training observations and adequately fits the other observations. This leads to different generalization error properties in the sensitive subgroup compared with the other observations.
\textbf{(Cause 3)} A more subtle issue, which is at the heart of \cite{Besse2020_BiasSurvey} is finally related to the influence of confounding variables: the prediction accuracy can be penalized in a subset of observations because they share common properties with another subset of observations having a different output. These common properties can be related to an explicit variable in the data or to unobserved  latent variables.
In neural-network based applications with images as inputs, the confounding variables will generally be a subset of pixels with unforeseen specific properties. This was made popular by the Husky or Wolf example \cite{RibeiroSIGKDD2016} where huskies are classified as wolves when they are represented in images with snow because the wolves of the training set are generally represented as surrounded by snow.

Tackling the algorithmic bias in machine learning is then an important and ambiguous task. This is particularly true when using deep neural-network models on images for which the decision rules are humanly impossible to interpret in the general case.
In this paper, we then propose a new solution to tackle the algorithm bias issue in neural-network based classification of image. The key advantages of our strategy are that it is Neural-Network model agnostic and that it scales particularly well to massive training sets of images.
Two variants of our method are presented, each of them being pertinent in specific contexts. The first variant helps the classifiers to predict similar frequencies of positive outputs in two groups of data (\textit{e.g.} in pictures representing males or females). It is then related to the notion of statistical parity. The second variant favor  similar error rates in the subgroups and is therefore related to the complementary notion of Equalized odds  \cite{hardt2016equality}.
This second variant is of particular interest in industrial frameworks, where a similar  accuracy has to be guaranteed in different contexts. Hence removing the bias is an important task when dealing with the generalization of predictions with different conditions modeled by the variable $S$. It appears to be an important tool for domain adaptation.

\section{Bibliography}\label{sec:biblio}

\subsection{Standard measures of the algorithmic bias}\label{ssec:MeasuringTheBias}

Different indices are commonly used to measure the algorithmic bias. We refer to  \cite{hardt2016equality,oneto2020fairness,DelBarrio2020ReviewFairnessML} for recent reviews of these measures.
In order to introduce them properly, we first denote $(x_i, s_i, y_i)_{i = 1,\dots, n}$  the training observations, where $x_i \in \mathbb{R}^p$ is an input image with $p$ pixels or voxels. Although we will treat 2D RGB images in our tests the $x_i$ may be in any dimension (2D, 3D, 2D+t, $\ldots$) and contain various amount of channels. In our paper, the output prediction $x_i \in \{0,1\}$ related to $x_i$ is supposed to be binary  for the sake of simplicity. We refer to \cite{hebert2018calibration,kearns2018preventing} for discussions about the multiple protective attributes and regression cases.
The protective variable $S \in \{0,1\}$ indicates whether observation $i$ is in a group which may be subject to algorithmic bias or not. It is common to use $s_i=0$ for the group which may be penalized and $s_i=1$ for the other data. The parameters $\theta$ of a binary classifier $g_{\theta}$ are  trained using the observations $(x_i, s_i, y_i)_{i = 1,\dots, n}$. The trained classifier is then used to predict the outputs $y_{i}$ of new input observations $x_{i}$, with $i>n$, so a prediction
$\hat{y}_{i} = g_{\theta}(x_{i}) \in \{0,1\}$.
Note that for each observation $i$, the neural network classifier returns in practice
a score $f_{\theta}(x_{i}) \in [0,1]$ which measures how  $g_{\theta}(x_{i})$ likely is to be equal to $1$. In general, $g_{\theta}(x_{i}) = \mathds{1}_{f_{\theta}(x_{i})>\frac{1}{2}}$.

The most standard measure of algorithmic bias is the so-called \emph{Statistical Parity}, which is often quantified in the fair learning literature using the Disparate Impact
(DI). This was been introduced in the US legislation in 1971\footnote{https://www.govinfo.gov/content/pkg/CFR-2017-title29-vol4/xml/CFR-2017-title29-vol4-part1607.xml} and measures how the variable $Y$ depends on $S$.

More specifically, $DI(Y,S)$ is defined as the ratio between $\mathbb{P}(Y=1|S=0)$ and $\mathbb{P}(Y=1|S=1)$, where we suppose that $S=0$ is  the group which may be discriminated with respect to $Y$.
The smaller this index, the stronger the discrimination over the group $S=0$.
A threshold $\tau_0=0.8$ is commonly used to judge whether the discrimination level of an algorithm is acceptable or not \cite{FeldmanSIGKDD2015,ZafarICWWW17,mercat2016discrimination,gordaliza2019obtaining}.
This  fairness criterion can be straightforwardly extended to the outcome of an algorithm by using  $\hat{Y}=g_{\theta}(X)$ instead of the true variable $Y$, and by making no hypothesis on the potentially discriminated group:
\begin{equation}\label{def:DIclassifier}
DI(g_{\theta},X,S)=\frac{\min \left[ \mathbb{P}(g_{\theta}(X)=1|S=s) \right]_{s \in \{0,1\}}}{\max \left[ \mathbb{P}(g_{\theta}(X)=1|S=s) \right]_{s \in \{0,1\}}} \,.
\end{equation}

A concern with the notion of statistical parity is that it implies that the decisions should be statistically the same for all subgroups of $S$, while in many situations, the end-user is more interested to detect whether the prediction algorithm has the same behavior for the observations out of each subgroup. In particular, the statistical parity does not take into account the false positive and false negative predictions, and more generally the prediction errors. As discussed in \cite{hardt2016equality}, the notions of equality of odds and opportunity additionally use the true predictions $Y$ and may then be more suitable than the statistical parity when a similar prediction accuracy (\textit{i.e.} $Acc = \mathbb{P}(\hat{Y}=Y)$) is desired in the subgroups $S=0$ and $S=1$.
We use the following notation to discuss these notions  \begin{equation}\label{def:EqOddsGeneralEq}
O_{s,y} = \mathbb{P} (\hat{Y}=1  |  S=s , Y=y ) \,.
\end{equation}
In the binary case, the classifier $g_{\theta}$ gives an \emph{equal opportunity} with respect to the protective attribute $S$ when $O_{0,1}$ equals $O_{1,1}$,
which means that the true positive prediction rate is the same in $S=0$ and $S=1$. Equalized odds with respect to $S$ are also satisfied if $O_{0,0}$ is additionally equal to $O_{1,0}$, which means that the false positive prediction rate is also the same  $S=0$ and $S=1$.
As the exact equality between two empirical probabilities does not necessarily make sense in practice, the predictions of a binary classifier will be considered as fair if $O_{0,y}$ reasonably close to $O_{1,y}$, for the true outputs $y=0$ and $y=1$.

We finally want to emphasize that a purely random binary classifier has a $DI$ close to $1$, and values of $O_{0,0}$ and $O_{0,1}$ similar to   $O_{1,0}$ and $O_{1,1}$, respectively. Its predictions are indeed independent of $S$. Despite of these nice properties, its accuracy is obviously almost equal to $0.5$, so such a classifier is  useless in practice. We then consider in this paper that a fair binary classifier has only an interest if its accuracy is reasonably close to $1$.

\subsection{Tackling the algorithmic bias}

In all generality, the notion of \textit{fairness} in Machine Learning in \cite{JohndrowAAS2019,pmlr-v97-mary19a,pmlr-v97-williamson19a,JiangUAI19}, consists in modeling  the \textit{algorithmic bias} as the independence (or the conditional independence) between the output $Y$ of an algorithm and a variable $S$ (or a group of variables).
Exhaustive bibliographies dealing with how to tackle this bias can be found in \cite{JiangUAI19,chouldechova2017fair,gordaliza2019obtaining,ZafarICWWW17,RothenhauslerArxiv2018,KusnerNIPS2018,hardt2016equality}.
Achieving this independence is indeed a difficult task which is obtained in the literature by favoring various fairness criteria that convey a part of the independence between $Y$ and $S$. In a sense, the algorithmic bias criteria used in the literature enable to weaken the notion of independence to achieve a good trade-off between fairness and prediction accuracy.
This idea was recently formalized as a \textit{price for fairness}  in \cite{gouic2020projection}.
Among the most popular fairness criteria, the Disparate Impact, the Equality of opportunity and the Equality of Odds \cite{hardt2016equality} presented in Section~\ref{ssec:MeasuringTheBias} are indirectly based on the  probability distribution of the outputs $f_{\theta}(X)$ with respect to a protected variable $S$. Their interpretation is clear and they are, in the authors opinion, powerful and straightforwardly interpretable tools to detect unfair decision rules.

An important concern with these measures is however that they are not smooth, which we believe is necessary to blend mathematical theory with algorithmic practice, in particular when using gold-standard gradient descent based optimization methods: For instance, if two observations $i$ and $j$ have similar non-binarized outputs $| f_{\theta}(x_i) - f_{\theta}(x_j) | < \epsilon$ (with $\epsilon$ small) but different binarized outputs
$\hat{y_i}=g_{\theta}(x_{i}) = \mathds{1}_{f_{\theta}(x_{i})>0.5}=0$ and $\hat{y_j}=g_{\theta}(x_{j}) = \mathds{1}_{f_{\theta}(x_{j})>0.5}=1$  these fairness criteria do not take into account the fact that the outputs $f_{\theta}$ are nearly the same. They won’t therefore directly favor observation $i$ instead of another observation to make the predictions more fair. As a result, this can make it hard to numerically find a good balance between fair and accurate predictions.

The covariance of the outputs $Y$ with respect to $S$ has also been investigated for instance in \cite{ZafarICWWW17} to directly measure linear independence. The variability of the loss function with respect to the protected variable has also been considered for similar purposes in \cite{pmlr-v97-williamson19a}. These two strategies have again these concerns dealing with the fact that they only measure the fairness based on $Y$. This clearly justifies for us the use of fairness measures taking into account the whole distribution, as the Wasserstein Distance, in particular when learning fair decision rules and not only when detecting unfair decision rules.
Another point view is given by considering the distance between conditional distributions. General divergences, Kullback-Leibler divergence or total variation have been considered  for Hilbert-Schmidt dependency in \cite{10.5555/3120007.3120011,komiyama2017twostage,Perez-SuayLMMGC17}.
Contrary to a transport based strategy like the Wasserstein distance, these measure are however not symmetric and cannot therefore be considered as metrics. Moreover the Wasserstein metric will make it possible to compare different measures, even if the intersection of their supports is null, which can be the case in real-life applications.

\subsection{Overview of the contribution}

We therefore propose to use the Wasserstein metric when training Neural Networks, which is smoothly defined and does not degenerate on empirical distributions given by finite samples. This distance has been already used in previous works dealing with fairness either to repair data or to build fair algorithms \cite{JiangUAI19,FeldmanSIGKDD2015}.  Wasserstein distance appears in this framework as a smooth criterion to assess the sensitivity w.r.t to the protected variable. As a matter of fact, it can be considered as a distance between the quantiles of score function of the predictor for the two groups (S=0 and S=1). The distance between the quantiles of these two distributions therefore acts as a level of fairness measuring whether the spread of the scores is homogeneously spread whatever the values of the protected attribute, hence acting as a  sensitivity index of the predicted values $f_{\theta}$ with respect to $S$.
Remark that, in \cite{JiangUAI19}, which is the closest work to our contribution, the authors specifically  used Wasserstein-1 to post-process the output of a model  to favor fair predictions. In our work, we instead use a more feasible criterion based on Wasserstein-2 regularization, as motivated in~\cite{gordaliza2019obtaining} to impose fairness constraints.
In some technical aspects, this also distinguishes our method from the optimization models used in Wasserstein GANS \cite{Arjovsky2017,Biau2021}, which are also based on Wasserstein-1 regularization.
Computing for a neural network the gradient of Wasserstein-2 distances with respect to machine learning model parameters is however an important technical lock to address, which is at the heart of Section~\ref{sec:methodo}.
We finally remark that relationships between Wasserstein distances and usual disparate impact are further discussed in \cite{DelBarrioII2018}.

We additionally focus on how to train neural network classifiers with fairness constraints for images since neural networks are  particularly flexible models and can treat huge volumes of data. Very little work has however been done so far to ensure fair decisions with neural networks \cite{NguyenAnnaArxiv2018,ManishaArxiv2018,RaffDSAA2018}. The existing literature also does not explicitly explain how to compute the gradients of the loss terms ensuring fair decisions. It additionally does not use Wasserstein-based regularization,  which was recently shown in  \cite{JiangUAI19} as an interesting solution for this purpose. In addition to the use Wasserstein-2 regularization instead of Wasserstein-1 regularization, our methodology differs from \cite{JiangUAI19} in different aspects.
We first make our predictions using neural-networks and not logistic-regression in order to train more flexible decision rules. By using the standard back-propagation formalism, we therefore do not directly compute the Wasserstein distance gradients with respect to the model parameters but with respect to the model outputs. As the Wasserstein distances are based on probability distribution, we then formalize and rigorously model a new interpretation of the impact of specific neural-network outputs on the continuous distribution of its outputs. We then show how to discretize this model with a numerical strategy that scales well to large datasets.
We finally explicitly develop how to handle the estimation of these gradients when using mini-batch training, which is extremely common for neural networks.

\section{Methodology}\label{sec:methodo}

\subsection{Main notations}\label{ssec:GenFormalism}

We recall that $(x_i, s_i, y_i)_{i = 1,\dots, n}$ are the training observations, where $x_i \in \mathbb{R}^p$ and $y_i \in \{0,1\}$ are the input and output observations, respectively.
Remark here that we denote by $x_i$, resp. $y_i$, i.i.d observations of the random variable $X$, resp. $Y$.
The protective variable $S \in \{0,1\}$ indicates whether observation $i$ is in a sensitive group ($s_i=0$) or not ($s_i=1$).
A classifier $\hat{g}_{\theta}$ with parameters $\theta$ is trained so that its predictions
$\hat{y_i}=g_{\theta}(x_{i}) \in \{0,1\}$ are, as often as possible, equal to the output observations $y_i$. Importantly, the binary predictions $g_{\theta}(x_{i})$ are defined as equal to
 $\mathds{1}_{f_{\theta}(x_{i})>0.5}$, where the continuous score $f_{\theta} \in [0,1]$ is the actual neural network output.
The problem can then be solved by minimizing a risk
\begin{equation}
\mathcal{R}(\theta):= \mathbb{E}[\ell (f_{\theta} (X), Y)] \,,
\end{equation}
or empirically $R(\theta) = \frac{1}{n} \sum_{i=1}^n \ell (f_{\theta} (x_i), y_i)$, where the loss function $\ell$ represents the price paid for inaccuracy of predictions
\cite{Bottou_SIAMrev2018}.

\subsection{Binary classification using Neural Networks}\label{ssec:prezNN}

The optimization of  the parameters $\theta$ over a compact set $\Theta$ is typically achieved using standard stochastic gradient descent \cite{Benveniste90,Bottou98} or its variants \cite{Duchi11,Kingma14} which require less computations than standard gradient descent and enable to explore more efficiently the parameters space.
At each iteration of the stochastic gradient descent, the average gradient of $R(\theta)$, and then $\mathcal{R}(\theta)$, is approximated on a subset  $B$ of several observations. This subset is denoted a mini-batch and  contains an amount of  $\#B$   observations.

Training neural-networks is specific in the sense that the parameters $\theta$ are indirectly optimized based on the derivatives of $R(\theta)$ with respect to network outputs $f_{\theta} (x_i)$, $i \in B$. In our tests, we defined $\ell (f_{\theta} (x_i), y_i)$ is equal to $\left( f_{\theta} (x_i)  - y_i \right)^2$. The empirical derivatives are then  computed using $2 \left( f_{\theta} (x_i) - y_i \right)$.
These derivatives are then back-propagated in the neural network and the parameters $\theta$ are updated using the stochastic gradient descent approach \cite{Rumelhart88,Bottou_SIAMrev2018}.
Many modern tools such as TensorFlow, Keras or PyTorch, make it simple to implement such training strategies based on automatic differentiation.
Note that although we used the mean squared error loss in our paper, the regularization method proposed in this paper can be directly used with other losses (\textit{e.g.} Binary Cross Entropy) as long as their derivative with respect to $f_{\theta} (x_i)$ can be computed.

\subsection{Wasserstein-2 based regularization to favor low discriminate impacts}\label{ssec:w2reg}

We denote by $\mu_{\theta,0}$ and $\mu_{\theta,1}$ the output distributions of  $f_{\theta}(X)$ for observations in the groups $S=0$ and $S=1$, respectively, and denote by $h_0$ and $h_1$ their densities.
Our regularization strategy consists in ensuring that the Wasserstein-2 distance (or Kantorovich-Rubinstein metric) between the  distributions of $\mu_{\theta,0}$ and $\mu_{\theta,1}$ remains small compared with $\mathcal{R}(\theta)$.
Their corresponding cumulative distribution functions are $\mathcal{H}_0$ and $\mathcal{H}_1$.
The Wasserstein-2 distance between the two conditional distributions is defined as
\begin{equation}\label{eq:w2regstar}
\mathcal{W}_2^2 (\mu_{\theta,0},\mu_{\theta,1}) = \int_0^1   \left( {\mathcal{H}_0}^{-1} (\tau) - {\mathcal{H}_1}^{-1} (\tau) \right)^2 d \tau \,.
\end{equation}
where ${\mathcal{H}_s}^{-1}$ is the inverse of the cumulative distribution function  $\mathcal{H}_s$.
Using the  distance Eq.~\eqref{eq:w2regstar} as a regularizer ensures that the distributions of $f_{\theta}(X)$ for the observations in the groups $S=0$ and $S=1$ remain reasonably close to each-other.
The training problem in a continuous setting is then:
\begin{equation}\label{eq:minimizedEnergy}
\hat{\theta} = \argmin_{\theta \in \Theta} \left\{ \mathcal{R}(\theta) + \lambda \mathcal{W}_2^2 (\mu_{\theta,0},\mu_{\theta,1}) \right\} \,,
\end{equation}
where $\lambda$ is a weight giving more or less influence to the regularization term compared with the prediction accuracy.

\subsection{Fast estimation of Wasserstein-2 pseudo-gradients in a batch}\label{ssec:QuickW2inBatch}

We have seen in Section~\ref{ssec:prezNN} that neural-network parameters are typically trained using the average gradient of the empirical risk $\mathcal{R}(\theta)$ with respect to all parameters on mini-batch observations.
Our goal here is to minimize Eq.~\eqref{eq:minimizedEnergy} when we observe empirical versions of the distributions $\mu_{\theta,0}$ and $\mu_{\theta,1}$ in the regularization term.
Standard automatic differentiation, which is ubiquitously used to train deep neural networks, cannot indeed be used
to minimize Eq.~\eqref{eq:minimizedEnergy} since the Wasserstein constraint Eq.~\eqref{eq:w2regstar} is applied on the distributions
$\mu_{\theta,0}$ and $\mu_{\theta,1}$ and not on a subset of observations.
In this section, we then explain how we compute, and which sense we give, to the gradient of Eq.~\eqref{eq:w2regstar} with respect to a finite subset of neural-network predictions  $f_{\theta}(x_i)$, $i \in B$.
To achieve this, we  model the regularization term of  Eq.~\eqref{eq:minimizedEnergy} as a function of the data distribution, denoted as $\mu_\theta$ and introduce a notion of derivative of  $\mu_\theta \mapsto \Phi(\mu_\theta)$ with respect to the distributional changes.
Discretization of our methodology is then developed in Section~\ref{ssec:discretecase}.

Note that similar arguments have for instance been used in \cite{Sommerfeld_and_Munk2018} in the discrete case or \cite{KitagawaJEMS2019} in the semi-discrete case. A similar notion of derivative in a geometric sense was also developed in the
Proposition 7.17. of \cite{Santambrogio2015}. None of these contributions however specifically modeled how to backpropagate the
impact of probability density variations.

\subsubsection{Modeling the impact of probability density variations on the transport cost}

In all generality, let $\mu,\nu\in \mathcal{P}(\mathbb{R})$ be two probability measures over $\mathbb{R}$. Consider as well  $(\alpha,\beta)\in\mathcal{M}(\mathbb{R})\times \mathcal{M}(\mathbb{R})$, where $\mathcal{M}(\mathbb{R})$ denotes the space of regular measures over $\mathbb{R}$. Inspired by \cite{Sommerfeld_JRSS2018,Tameling_AAP2019},  we can remark that   the transport cost is  G\^ateaux differentiable in the direction $(\alpha,\beta)$, with derivative $D\mathcal{W}^2_2(\mu,\nu)(\alpha,\beta)$ if the limit
\begin{equation}\label{gateaux}
   D\mathcal{W}^2_2(\mu,\nu)(\alpha,\beta):= \lim_{t \searrow 0}\frac{\mathcal{W}^2_2(\mu+t\alpha,\nu+t\beta)-\mathcal{W}^2_2(\mu,\nu)}{t}
\end{equation}
exits and is finite. Note that the value $\mathcal{W}^2_2(\mu+t\alpha,\nu+t\beta)$ only makes sense when $\alpha(\mathbb{R})=\beta(\mathbb{R})$, where
$\alpha(\mathbb{R})$ and $\beta(\mathbb{R})$ denote the total masses of the measures and are not necessarily positive (see \textit{e.g.} \cite{RudinBook1987}, chapter 6).  Otherwise, the transport would not be  defined any longer. Remark that $\mu+t\alpha$ and $\nu+t\beta$ should not take negate values too, as they represent probability measures.
\begin{proposition}\label{prop:GateauxDiffTransportCost}
Let $\mu,\nu\in \mathcal{P}(\mathbb{R})$ with density w.r.t  Lebesgue measure and supported on respective  connected sets and $(\alpha,\beta)\in\mathcal{M}(\mathbb{R})\times \mathcal{M}(\mathbb{R})$ be such that
\begin{enumerate}
   \item $\alpha(\mathbb{R})=\beta(\mathbb{R})$,
    \item $\operatorname{Supp}(\alpha)$ and $\operatorname{Supp}(\beta)$ are both compact sets, \item $\operatorname{Supp}(\alpha)\subset \operatorname{Supp}(\mu)$ and  $\operatorname{Supp}(\beta)\subset \operatorname{Supp}(\nu)$.
\end{enumerate}
 Then the  Transport cost is G\^ateaux differentiable in the direction $(\alpha,\beta)$ and it satisfies
 \begin{equation*}
    D\mathcal{W}^2_2(\mu,\nu)(\alpha,\beta)=2\int \int_{0}^{x}(t-F^{-1}_{\nu}(F_{\mu}(t)))dtd\alpha(x)
    +2\int \int_{0}^{x}(t-F^{-1}_{\mu}(F_{\nu}(t)))dtd\beta(x) \,,
 \end{equation*}
where $F_{\mu}$ and $F_{\nu}$ are the cumulative distribution functions of $\mu$ and $\nu$.
\end{proposition}
A proof of Proposition~\ref{prop:GateauxDiffTransportCost} is given in Appendix~\ref{sec:ProofPropGradW2obs}.

In our application, our goal is to  quantify the impact of variations of $f_{\theta}(x_i)$ on either $\mu_{\theta,0}$ or $\mu_{\theta,1}$, depending on the label of $s_i$. Suppose now that we are in the  case  $s_i=0$.
In order to later model a perturbation of only one of the two distributions, we consider now  the direction $(\alpha,0)$, where $\alpha(\mathbb{R})=0$,  so that $\mu+t \alpha$  remains a probability on $\mathbb{R}$.
Let $u$ be the density of $\mu$. As $d\mu(t) = u(t)d(t)$,  we can write that
\begin{equation}\label{eq:generalcase}
\int \int_{0}^{x}     (t-F^{-1}_{\nu}(F_{\mu}(t)))dtd\alpha(x)
                    =\int \int_{0}^{x}\frac{t-F^{-1}_{\nu}(F_{\mu}(t))}{u(t)}d\mu(t)d\alpha(x) \,,
\end{equation}
in the case where the right side is finite. This will be the case in our specific application, as the predictions $f_{\theta}(x_i)$ are supposed to be finite.\\

\subsubsection{Influence of the perturbation of an output $f_\theta(x_i)$ on the Wasserstein-2 distance}\label{ssec:continuouscase}

We now specify Eq.~\eqref{eq:generalcase} to quantify  the influence of a perturbation of the output prediction $f_\theta(x_i)$, which is the very first step of a back-propagation algorithm when training a neural network.
To model model  a local perturbation  around $f_\theta(x_i)$ using a  measure, we  first consider $\epsilon>0$ and set
$\alpha_{\epsilon}^i=\frac{1}{2\epsilon}\mathbb{I}_{[0,f(x_i)+\epsilon]}-\frac{1}{2\epsilon}\mathbb{I}_{[f(x_i)-\epsilon,0[}$.
Hence $\alpha_{\epsilon}^i$ is supported in $[f_{\theta}(x_i)-\epsilon,f_{\theta}(x_i)+\epsilon]$, such that $\alpha_{\epsilon}^i(\mathbb{R})=0$ and is anti-symmetric  with respect to $f_{\theta}(x_i)$. We then have
\begin{equation}\label{eq:specificcase}
\int \int_{0}^{x}\frac{t-F^{-1}_{\nu}(F_{\mu}(t))}{u(t)}d\mu(t)d \alpha_{\epsilon}^i (x) =
\int_{f_{\theta}(x_i)-\epsilon}^{f_{\theta}(x_i)+\epsilon} \int_{0}^{x}\frac{t-F^{-1}_{\nu}(F_{\mu}(t))}{u(t)}d\mu(t)d \alpha_{\epsilon}^i (x)
\end{equation}
This equation can be re-written as:
\begin{equation*}
\frac{1}{2 \epsilon}
  \left[
    \int_{f_{\theta}(x_i)-\epsilon}^{f_{\theta}(x_i)+\epsilon}
          \underbrace{\left(\int_{0}^{x}\frac{t-F^{-1}_{\nu}(F_{\mu}(t))}{u(t)}d\mu(t)\right)}_{\Phi(x)}
    2 \epsilon d \alpha_{\epsilon}^i (x)
  \right] \,,
\end{equation*}
which is equal to
\begin{equation*}
\frac{1}{2 \epsilon}
  \left[
    \int_{0}^{f_{\theta}(x_i)+\epsilon} \Phi(x) d x -
    \int_{f_{\theta}(x_i)-\epsilon}^{0} \Phi(x) d x
  \right] \,,
\end{equation*}
and tends towards $\Phi'(f_{\theta}(x_i))$ when $\epsilon \rightarrow 0$. In this case, Eq.~\eqref{eq:specificcase} is then approximated by:
\begin{equation*}
\frac{f_{\theta}(x_i)-F^{-1}_{\nu}(F_{\mu}(f_{\theta}(x_i)))}{u(f_{\theta}(x_i))} \,.
\end{equation*}
We then have the following proposition:
\begin{proposition}
Let $\mu=\mu_{\theta,0}$ and $\nu=\mu_{\theta,1}$, then we have that
 \begin{equation}\label{eq:specificcase2}
    \lim_{\epsilon \searrow  0}D\mathcal{W}^2_2(\mu,\nu)(\alpha_{\epsilon}^i,0)=
\frac{f_{\theta}(x_i)-\mathcal{H}^{-1}_{1}(\mathcal{H}_{0}(f_{\theta}(x_i)))}{\mathcal{H}_{0}'(f_{\theta}(x_i))}
 \end{equation}
\end{proposition}
This represents the influence of a perturbation made by the output prediction $f_\theta(x_i)$ if $s_i=0$. Note that the cumulative distribution functions $\mathcal{H}_{0}$ and $\mathcal{H}_{1}$ will straightforwardly be swapped in this equation if $s_i=1$.

\subsubsection{Impact of several observations in a mini-batch $B$}

The following lemma allows to extend the previous result to the mini-batch case.

\begin{lemma} \label{gat:los}
let $\mu\in \mathcal{P}(\mathbb{R})$ be a probability measure and $l$ be a real measurable function, then the function
\begin{align*}
    L: \mathcal{P}(\mathbb{R}):&\longrightarrow \mathbb{R}\\
    \mu&\mapsto \int l d \mu
\end{align*}
is G\^ateaux differentiable for every direction $\beta\in \mathcal{M}(\mathbb{R})$ such that $\int l d \beta <\infty $ with derivative
\begin{align*}
   DL(\mu;\beta)= \int l d \beta.
\end{align*}
\end{lemma}
A proof of Lemma~\ref{gat:los} is given in Appendix~\ref{sec:ProofPropGradW2obs}.
If $\mu_\theta$ depends on the protected attribute $S$, it is therefore easy to see that
$$L(\mu)=L(\mu_{\theta,0})P(S=0)+ L(\mu_{\theta,1})P(S=1),$$
then we can use the previous result with the direction $\alpha^{i}_{\epsilon}$ defined before and deduce that

\begin{align*}
DL(\mu_\theta;\alpha^{i}_{\epsilon})&= \frac{1}{2\epsilon}\left( l(f_\theta(x_i)+\epsilon)-l(f_\theta(x_i)-\epsilon) \right) \\
DL(\mu_\theta;\alpha^{i}_{\epsilon})&  \xrightarrow[\epsilon\rightarrow 0]{} (\nabla l)(f(x_i)).
\end{align*}

As a consequence, we can define the mean influence of the batch $B$ as the mean of the previous limit, which yields:
$$
   \frac{1}{\# B}\sum_{i \in B} (\nabla l)(f(x_i)) \,,
$$

Together with Proposition 1, this proves the G\^ateaux differentiability of  \[\mu_\theta \longmapsto  \lambda \mathcal{W}_2^2 (\mu_{\theta,0},\mu_{\theta,1}) .\]
when using batch training.

\subsubsection{Influence of an output $f_{\theta}(x_i)$ in the discrete case}\label{ssec:discretecase}

\begin{figure}[h]
  \centering
  \includegraphics[width=0.50\linewidth]{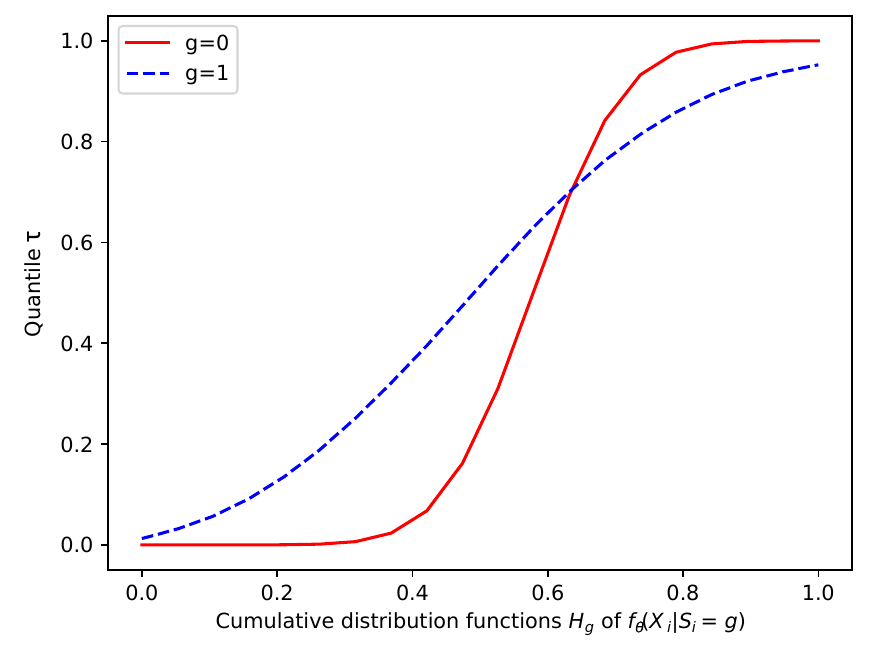}
  \caption{Cumulative distributions $H_0$ and $H_1$ of simulated predictions $f_{\theta}(x_i)|s_i=s$ for the two groups $s=0$ and $s=1$.}
  \label{fig:cumDistribs}
\end{figure}

We now extend the methodology of Section~\ref{ssec:continuouscase} to the discrete case, where the number of observations $f_{\theta}(x_i)$ is finite and the cumulative distributions are only known on discrete grids of values.
We first denote $\mu_{\theta,s}^n$ the approximation of $\mu_{\theta,s}$ made using $n$ observations.
For the group $S=s$, we denote $H_s(\eta)$ the discrete approximation of $\mathcal{H}_s(\eta)$. It represents the empirical portion of predictions $f_{\theta}(X)$ in group $s$ having a value lower or equal to $\eta \in [0,1]$, \textit{i.e.}
$H_s (\eta) = n_s^{-1} \sum_{i=1}^n \mathbbm{1}_{f_{\theta}(x_i)<\eta \textrm{ and } s_i=s}$, where $n_s$ is the number of observations in group $s$. As illustrated in Fig.~\ref{fig:cumDistribs}, the $H_s$ are pre-computed on discrete grids of values:
\begin{equation}\label{eq:PbDiscretization}
\eta^j = \min_i(f_{\theta}(x_i)) + j \Delta_{\eta} \,,\, j=1, \ldots, J_{\eta} \,,
\end{equation}
where $\Delta_{\eta} = J_{\eta}^{-1} (\max_i(f_{\theta}(x_i)) - \min_i(f_{\theta}(x_i)) )$ and $J_{\eta}$ is the number of discretization steps. We denote $H_s^j = H_s (\eta^j)$.
The regularization function Eq.~\eqref{eq:w2regstar} is then estimated by using the empirical distributions
\begin{equation}\label{eq:w2reg}
\mathcal{W}_2^2 (\mu^n_{\theta,0},\mu^n_{\theta,1}) = \int_0^1   \left( H_0^{-1} (\tau) - H_1^{-1} (\tau) \right)^2 d \tau \,,
\end{equation}
where $H_s^{-1}$ is the \textit{inverse} of the empirical function   $H_s$, \textit{i.e.}
$H_s^{-1}(\tau)$ is the $\tau$'th quantile of the observed values $f_{\theta}(x_i)$ for $s_i=s$. As the functions $H_s (\eta)$ are discontinuous and are only known for the values $\eta^j$ of Eq.~\eqref{eq:PbDiscretization}, they are inverted using linear interpolations of the $H_s^j$.
The integration of Eq.~\eqref{eq:w2reg} is finally performed on a discrete grid of $J_{\tau}$ values $\tau$ with a  discretization step-size $\Delta_{\tau} = 1/J_{\tau}$. We denote $W_2^2 (\mu^n_{\theta,0},\mu^n_{\theta,1})$ this approximation.

\begin{figure}[h]
  \centering
  \includegraphics[width=0.50\linewidth]{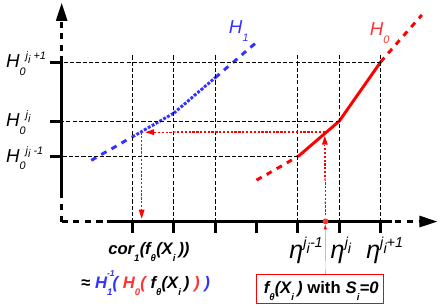}
  \caption{Notations used to efficiently study the influence of a change in the outputs $f_{\theta}(x_i)$ on the Wasserstein-2 distances between the discrete cumulative distributions $H_0$ and $H_1$.}
  \label{fig:DetailedCumDistribs}
\end{figure}

In this discrete context, the derivative of $H_s$ will be approximated by $\left(H_s^{j_i+1}-H_s^{j_i}\right)/ \Delta_{\tau}$ using finite differences, where $s_i=s$ and $j_i$ is defined such that $\eta^{j_i} \leq f_{\theta}(x_i) < \eta^{j_i+1}$. Interestingly, this discrete derivative will never be equal to zero for a value of $f_{\theta}(x_i)$ as long as this prediction was used to compute the $H_s^j$.
As illustrated Fig.~\ref{fig:DetailedCumDistribs}, we also denote $cor_s(f_{\theta}(x_i)) $ the value corresponding to $f_{\theta}(x_i)$ in the group $S=|1-s|$ when integrating Eq.~\eqref{eq:w2reg}, \textit{i.e.}
$cor_s(f_{\theta}(x_i)) = H_{s}^{-1}(H_{|1-s|}(f_{\theta}(x_i)))$.

Then the empirical counterpart of Eq.~\eqref{eq:specificcase2} is
\begin{equation}\label{eq:gradS0S1}
   \frac{\Delta_{\tau}({f_{\theta}(x_i)-cor_{|1-s_i|}(f_{\theta}(x_i))})}{(H_{s_i}^{j_i+1}-H_{s_i}^{j_i})} \,,
\end{equation}
for $s_i \in \{0,1\}$.
As a consequence, the variation of $\mu_\theta \mapsto  W_2^2 (\mu_{\theta,0}^n,\mu_{\theta,1}^n)$ due to the prediction $f_{\theta}(x_i)$, with $i \in B$, can be back-propagated using the expression
\begin{equation}\label{eq:finalW2gradApprox}
 \displaystyle{
\Delta_{\tau} \left[
 \mathds{1}_{s_i=0}
 \frac{     f_{\theta}(x_i) - cor_1(f_{\theta}(x_i))   }{ n_0 \left(H_{0}^{j_i+1} - H_{0}^{j_i} \right)}
    -  \mathds{1}_{s_i=1}
      \frac{     cor_0(f_{\theta}(x_i)) - f_{\theta}(x_i) }{ n_1 \left(H_{1}^{j_i+1} - H_{1}^{j_i} \right)}
     \right] \,,
   }
   \end{equation}
where we added the terms $n_0$ and $n_1$ to Eq.~\eqref{eq:gradS0S1} with $s_i=0$ and $s_i=1$ in order to have a balanced influence of the classes $S_0$ and $S_1$ when computing the average gradients in a mini-batch.

\subsubsection{Algorithmic pertinence of Eq.~\eqref{eq:finalW2gradApprox} on large training sets}\label{sssec:algorithmicPertinence}

For a group $s$, we denoted the discrete  cumulative distribution $H_s = \{H_s^j\}_{j=1, \ldots, J}$ such as $H_s^j$  is equal to $n_s^{-1} \sum_{i=1}^n \mathbbm{1}_{f_{\theta}(x_i)<\eta^j \textrm{ and } s_i=s}$, where  $n_s$ is the number of observations in group $s$. These quantiles can be computed once for all before estimating all derivatives in a batch $B$. Computing $H_0$ and $H_1$ using all observations has an algorithmic cost $o(n)$. They can also   be computed on reasonably large random subsamples of $n_{sub}$  observations if $n$ is extremely large, so the algorithmic cost can be reduced to  $o(n_{sub})$. This makes this pre-computation tractable in all cases.
For a given observation $i \in B$, we also denoted $j_{i}$ the index such that $\eta^{j_{i}} \leq f_{\theta}(x_i) < \eta^{j_{i}+1}$. This index can be found for a very reasonable  algorithmic cost of $o(\log_2(J))$ using a divide and conquer approach.
Finally that the $cor_s$ can be computed using linear interpolation, which is again algorithmically  reasonable.

As a consequence, Eq.~\eqref{eq:finalW2gradApprox} can be efficiently computed in a batch as the $\eta^j$ are known, the  $H_{{s_i}}^{j}$ are pre-computed once for all in each batch,  the search for the $j_{i}$ is algorithmically reasonable, and the linear interpolations to compute the $cor_s(f_{\theta}(x_i))$ is algorithmically cheap.

Remark that in order to make fine and stable the empirical estimation of $\mu_{\theta,0}$ and $\mu_{\theta,1}$,
more observations than only those of $B$, but still far less than all $n$ observations, can be used. For instance, in the results obtained Section~\ref{ssec:CelebAdataset}, the batches contained 2000 observations and the empirical output distributions were computed using 8000 additional observations. Importantly, it was not required to store any information for the automatic differentiation when computing these additional observations (specifically, the \textit{torch.no\_grad} mode was used in \textit{PyTorch}). This strongly limited the memory footprint of these predictions.

\subsection{Favoring similar error rates}\label{sssec:sim_error_rates}

\subsubsection{Introduction}

An important variant of Sections~\ref{ssec:w2reg} and  \ref{ssec:QuickW2inBatch} consists in favoring decision rules leading to similar true positive and/or true negative rates for $S=1$ and $S=0$, and not only a similar portion of true decisions. This is related to the notion of equality of opportunities \cite{hardt2016equality}, which was discussed in Section~\ref{ssec:MeasuringTheBias}.

In order to favor similar true positive predictions only, the method of
Section~\ref{ssec:QuickW2inBatch} can be simply extended by only computing the cumulative distributions on the observations for which the true prediction $y_i$ is $1$. The same idea holds for the true negative predictions by using the observations for which $y_i=0$. Favoring a similar error rate in the groups $S=0$ and $S=1$ is slightly more complex as described hereafter.

We first use a non-binary definition of  equalized odds conditions
(presented in Section~\ref{ssec:MeasuringTheBias}), as in Eq.~\eqref{def:DIclassifier}, and merge these two conditions on the observations $Y=1$ and $Y=0$. To do so, we will use the following index denoted by the \textit{Disparate Mean Squared Error} (DMSE):
\begin{equation}\label{def:DEclassifier}
DMSE(g_{\theta},X,S)=\frac{\min \left[ MSE(g_{\theta},X,Y | S=s) \right]_{s \in \{0,1\}}}{\max \left[ MSE(g_{\theta},X,Y | S=s) \right]_{s \in \{0,1\}}} \,,
\end{equation}
where the mean squared error of the  classifier is empirically defined as:
\begin{eqnarray}\label{def:groupMSE}
MSE(g_{\theta},X,Y| S=s) &=& \frac{1}{n_s} \sum_{i=1\,,\, s_i=s}^{n} (g_{\theta}(x_i)-y_i)^2   \nonumber\\
                        &=& \frac{1}{n_s} \sum_{i=1\,,\, s_i=s}^{n} \mathds{1}_{y_i=g_{\theta}(x_i)} \,,
\end{eqnarray}
where $n_s=\sum_{i=1}^{n} \mathds{1}_{s_i=s}$ and we recall that $g_{\theta}(X_{i})=\mathds{1}_{f_{\theta}(X_{i})>0.5}$.
It is straightforward to show that this index is equal to $1$ if the equalized odds conditions are respected and that the more disproportionate the error rate between $S=0$ and $S=1$ the closer this index to $0$.

Motivated by the same optimization concerns as in Section~\ref{ssec:w2reg}, we first propose to use a slightly modified definition of $MSE$ which is continuous with respect to the parameters $\theta$
\begin{equation}\label{def:continuous_groupMSE}
cMSE(f_{\theta},X,Y| S=s) = \frac{1}{n_s} \sum_{i=1\,,\, s_i=s}^{n} (f_{\theta}(x_i) - y_i)^2 \,,
\end{equation}
and to use this definition in the DMSE index, Eq.~\eqref{def:DEclassifier}.
Our main contribution is then to propose to minimize this index by using the Wasserstein-2 distance between the densities of squared errors obtained in the groups $S=0$ and $S=1$.

\subsubsection{Back-propagating  the impact of an observation}\label{ssec:err_pseudoGrads}

We denote by $\tilde{\mu}_{\theta,0}$ and $\tilde{\mu}_{\theta,0}$ the densities of  the squared error rates $(f_{\theta} (X)-Y)^2$ in the groups $S=0$ and $S=1$, respectively, and
$\tilde{\mu}_{\theta,0}^n$ and $\tilde{\mu}_{\theta,0}^n$ their discrete counterparts. For a group $s$, we then denote the discrete  cumulative distribution $\tilde{H}_s = \{\tilde{H}_s^j\}_{j=1, \ldots, J}$ such as
\begin{equation}\label{eq:def_Htilde}
\tilde{H}_s^j
= n_s^{-1} \sum_{i=1\,,\,s_i=s}^n \mathbbm{1}_{(f_{\theta} (x_i)-y_i)^2<\eta^j  } \,,
\end{equation}
where $n_s$ is the number of observations in group $s$, and the $\{\eta^j\}_{j=1, \ldots, J}$ are regularly sampled thresholds between $0$ and $1$ (as in Section~\ref{ssec:QuickW2inBatch}).
Instead of solving Eq.~\eqref{eq:minimizedEnergy}, we solve here
\begin{equation}\label{eq:minimizedEnergyErr}
\hat{\theta} = \argmin_{\theta \in \Theta} \left\{ R(\theta) + \lambda W_2^2 (\tilde{\mu}^n_{\theta,0},\tilde{\mu}^n_{\theta,1}) \right\} \,.
\end{equation}

We develop in Appendix~\ref{sec:ProofgradW2squaredErrors} how to back-propagate efficiently the impact of an observation  $f_{\theta} (x_i)$ on the regularization term
$W_2^2 (\tilde{\mu}^n_{\theta,0},\tilde{\mu}^n_{\theta,1})$.
This leads to the fact that a variation of $\tilde{\mu}_\theta \mapsto  W_2^2 (\tilde{\mu}_{\theta,0},\tilde{\mu}_{\theta,1})$ modeling this impact can be back-propagated using the expression
\begin{align}\label{eq:finalW2gradApprox_err}
&\displaystyle{
2\Delta_{\tau}
\left[
 \mathds{1}_{s_i=0}
 \frac{    (f_{\theta}(x_i)-y_i)^2 - cor_1 \left((f_{\theta}(x_i)-y_i)^2\right)   }{n_0 \left(\tilde{H}_{0}^{j_i+1} - \tilde{H}_{0}^{j_i}\right) \left(f_{\theta}(x_i)-y_i\right)^{-1}}
 \right. } \nonumber  \\
&\displaystyle{
\left.
  -  \mathds{1}_{s_i=1}
\frac{    cor_0 \left((f_{\theta}(x_i)-y_i)^2 \right) - (f_{\theta}(x_i)-y_i)^2 }{n_1 \left( \tilde{H}_{1}^{j_i+1} - \tilde{H}_{1}^{j_i}\right) \left(f_{\theta}(x_i)-y_i\right)^{-1} }
\right] \,,
}
\end{align}
where the $cor_s$ have the same meaning as in Eq.~\eqref{eq:finalW2gradApprox} but are computed on the $\tilde{H}_s$ instead of the  $H_s$.
Computing this equation has a very similar algorithmic cost as the estimation step of Eq.~\eqref{eq:finalW2gradApprox}.
The Wasserstein-2 regularization is therefore computationally reasonable on the squared error.
Remark that other extensions of our method are possible. In particular, we extend our method in Section~\ref{sec:ExtensionsQuickW2inBatch} to the Wasserstein-1 penalty measure and the Logistic regression case.

\subsection{Training procedure}

\begin{algorithm}{ht}
\caption{Batch training procedure for neural-networks with Wasserstein-2 regularization}
\label{alg:BTP_NN_W2}
\begin{algorithmic}[1]
\REQUIRE Weight $\lambda$ and the training observations $(x_i,s_i,y_i)_{i=1,\ldots,n}$, where $x_i\in \mathbb{R}^p$, $s_i \in \{0,1\}$ and $y_i \in \{0,1\}$.
\REQUIRE Neural network $f_{\theta}$ with initialized parameters $\theta$.
\FOR{$e$ in Epochs}
\FOR{$b$ in Batches}
\STATE Pre-compute $H_0$ and $H_1$.
\STATE Draw the batch observations $B$.
\STATE Compute the $f_{\theta}(x_i)$, $i\in B$.
\STATE Compute the standard loss derivatives.
\STATE Compute the impact of the mini-batch outputs $f_{\theta}(x_i)$ on $W_2^2 (\mu^n_{\theta,0},\mu^n_{\theta,1})$ using Eq.~\eqref{eq:finalW2gradApprox}.
\STATE Backpropagate the impact of all $f_{\theta}(x_i)$, $i \in B$.
\STATE Update the parameters $\theta$.
\ENDFOR
\ENDFOR
\RETURN Trained neural network $f_{\theta}$.
\end{algorithmic}
\end{algorithm}

Our batch training procedure is summarized Alg.~\ref{alg:BTP_NN_W2} for the disparate impact case, \textit{i.e.} to keep similar output distributions $Y$ in the groups $S=0$ and $S=1$.
The procedure is the same for the squared error case, except that the $\{H_s,\mu_{\theta,s}\}$ are replaced by the $\{\tilde{H}_s,\tilde{\mu}_{\theta,s}\}$ (see Eq.~\eqref{eq:def_Htilde}) and Eq.~\eqref{eq:finalW2gradApprox} is replaced by  Eq.~\eqref{eq:finalW2gradApprox_err}.
It is important to remark that we implemented this training procedure in \emph{PyTorch}\footnote{https://pytorch.org/} by only writing a specific \emph{autograd.Function} for our regularized loss term.

\section{Results}

We assess in this section different aspects of the proposed method. In Section~\ref{ssec:Adultdataset}, we compare it to two other methods on the   \textit{Adult Census} dataset. Although this is a tabular dataset, it has become the gold standard dataset to assess the level of fairness of new  classification strategies. This justifies its use in the beginning of the results section. We then evaluate in Section~\ref{ssec:MNISTdataset} the influence of the weight $\lambda$ on the other standard \textit{MNIST} image dataset. In order to discuss in depth the results of this section, we simulated two kinds of bias in the training dataset with a controlled level of bias. We finally compare in Section~\ref{ssec:CelebAdataset} the two proposed regularization alternatives on the large \textit{CelebA} image dataset. In this section, we denote \textit{Reg. Prediction} the strategy of Section~\ref{ssec:w2reg} which favor similar predictions in $S=0$ and $S=1$, and we denote \textit{Reg. Error} the strategy of Section~\ref{sssec:sim_error_rates} which favor similar predictions errors in the two groups.

\subsection{Adult census dataset}\label{ssec:Adultdataset}

\subsubsection{Dataset}

In order compare the proposed method with different alternatives, we used the \textit{Adult Census} dataset\footnote{https://archive.ics.uci.edu/ml/datasets/adult}.
  It  contains $n=45222$ subjects and $p=14$ input variables. The binary output variable $Y$   indicates whether a subject's \textit{incomes} is above (positive class, so $Y=1$) or below (negative class, so $Y=0$) 50K USD. We also consider variable \textit{Gender} as sensitive. Note that the authors have extensively studied this dataset in \cite{Besse2020_BiasSurvey}. After discussing that the training set clearly contains more males than females with $Y=1$, they have made clear that naive bias correction techniques are inefficient to train decision rules leading to similar rates of predictions with $\hat{Y}=1$. This is because of the strong influence of confounding variables, which corresponds to the \textit{Cause 3} of algorithmic bias  in Section~\ref{ssec:AlgorithmicBiasMachineLearning}. Note that as discussed in \cite{Besse2020_BiasSurvey}, the training dataset also contains about two times more males than females and that the males have more frequently $Y=1$ than the females, leading potentially to the \textit{Cause 2} of bias.

\subsubsection{Experimental protocol}

The goal in these tests is to reach as much as possible a good balance between the prediction accuracy and the statistical parity of positive outputs w.r.t the gender.
We then denote \textbf{NNrW} and \textbf{NN} neural-network based classification strategies with and without the regularization method of Section~\ref{ssec:w2reg}. We first compared these neural-network classifiers to a logistic regression classifier with Wasserstein-1 regularization (\textbf{LRrW}), as in \cite{JiangUAI19}. Specifically, we  mimicked \cite{JiangUAI19} by using the technique of Appendix~\ref{sssec:LR_alternative} with the regularization of Appendix~\ref{sssec:W1_alternative}.
We also tested different variants of the reference method of \cite{ZafarICWWW17}, with constraints that explicitly favor high discriminate impacts (\textbf{ZFA}) or low rates of false negative predictions (\textbf{ZFN}). We finally used the non-regularized Logistic-Regression of Scikit-Learn\footnote{https://scikit-learn.org/stable/modules/linear\_model.html} (\textbf{LR}), which is the baseline method of \cite{ZafarICWWW17}.

Default parameters were used for the reference methods of LR,  ZFA and ZFN and 300 iterations were used for the gradient descents LRrW.
The strategies NNrW, and NN use the same elementary network architecture. It is  made of  three fully-connected hidden layers with Rectified Linear Units (ReLU) activation functions. A sigmoid activation function is then used in the output layer to allow binary classification. Optimization was made using Adam method \cite{Kingma14} with the default parameters on PyTorch. An amount of 100 epochs and a batch size of 50 observations were used.
The method of Appendix~\ref{sssec:autoTuneLambda} was also used to tune $\lambda$ in  NNrW.
Finally, similar pre-treatment as those of \cite{Besse2020_BiasSurvey} were made on the data to transform the categorical variables into quantitative variables.

All scores are given on test data after randomly splitting all available data into training and test datasets, with 75\% and 25\% of the observations each, respectively.
To measure the classification quality and fairness we considered the accuracy of the predictions (formally defined Section~\ref{ssec:MeasuringTheBias}), the disparate impact Eq.~\eqref{def:DIclassifier} and the empirical probabilities Eq.~\eqref{def:EqOddsGeneralEq}, which were derived from the equalized odds principle of \cite{hardt2016equality}  as $O_{s,y} = \mathbb{P} (\hat{Y}=1  |  S=s , Y=y )$.

\subsubsection{Results}

\begin{table}
  \caption{Adult dataset with sensitive variable \textit{Gender}}
  \label{tab:ResACsex}
\begin{center}
\begin{tabular}{|c||c||c||c|c|c|c|c|c|c|}
   \hline
      &Acc  &DI   &$O_{1,0}$  &$O_{0,0}$ &$O_{1,1}$ &$O_{0,1}$  \\ \hline \hline
LR    &0.84 &0.47 &0.18        &0.05       &0.71       &0.80 \\ \hline
NN    &0.82 &0.37 &0.20        &0.05       &0.73       &0.64 \\  \hline \hline
ZFN    &0.79 &0.64 &0.10        &0.07       &0.42       &0.43 \\ \hline
ZFA    &0.66 &0.95 &0.39        &0.42       &0.81       &0.83 \\ \hline \hline
LRrW    &0.65 &0.94 &0.40        &0.44       &0.85       &0.86 \\  \hline
NNrW    &0.78 &0.68 &0.25        &0.20       &0.85       &0.83 \\  \hline
\end{tabular}
\end{center}
\end{table}

Results are given in Table~\ref{tab:ResACsex}.
The results of LRrW are first similar to those of ZFA and can be considered as the most fair of all, both from the statistical parity ($DI\approx 1$) and the equalized odds ($O_{1,0} \approx O_{0,0}$ and $O_{1,1} \approx O_{0,1}$) perspectives. Their accuracy is however clearly lower than $0.8$, so they cannot be considered as reasonably accurate. The baseline methods LR and NN, which have no fairness constraints, are now the most accurate ones with $Acc>0.82$. Their DI is however particularly low as they favor much more false positive predictions in $S=1$ than $S=0$ \textit{i.e.} $O_{1,0} \gg O_{0,0}$. The methods leading to a good balance between accuracy and fairness are finally ZFN and NNrW. Although ZFA predicts well the negative outputs ($O_{1,0}$ and $O_{0,0}$ lower than $0.11$) it does not capture properly how to predict true positive
outputs ($O_{1,1}$ and $O_{0,1}$ lower than $0.44$, meaning more than 56\% error). This can be explained by the fact that only 24\% of the training observations have positive outputs. For a similar accuracy and DI, the strategy NNrW appears as more balanced when comparing to the true and false positive rates. The true positive predictions ($O_{1,1}$ and $O_{0,1}$) have less than 17\% of errors and the false positive predictions ($1-O_{1,0}$ and $1-O_{0,0}$) have less than 26\% of errors. The NNrW therefore appears as the one leading to the best balance between fairness and accuracy.

\subsection{MNIST dataset}\label{ssec:MNISTdataset}

\subsubsection{Unbiased data}\label{ssec:ExpProUnbiased}

\begin{figure*}[h]
  \centering
  \includegraphics[width=0.99\linewidth]{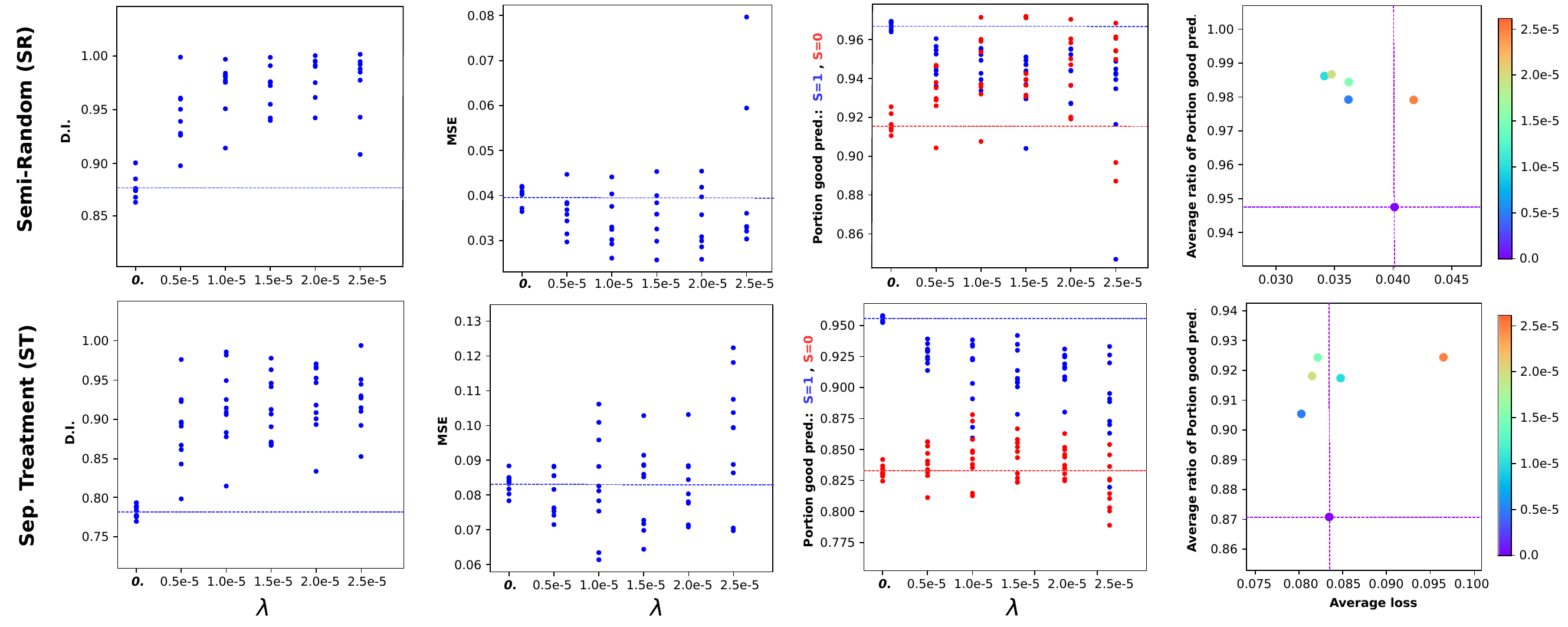}
  \caption{Results obtained on the MNIST dataset by following the SR and ST experimental protocols. Both for SR and ST, an amount of 9 classifiers where trained with $\lambda=0$ (no regularization) and     $\lambda = \{0.5, 1.0,1.5, 2.0,2.5\}e-5$. We represent, the disparate impact (DI), Mean Squared Error (MSE), the portion of good predictions in the groups $S=0$ and $S=1$  (GP0 and GP1) obtained  on the test set  using each trained classifier, as well as the average ratios GP0/GP1. The dashed lines in each subfigure show the average scores obtained with no regularization.
  }
  \label{fig:ResMNIST1}
\end{figure*}

%RESULTS IN: /home/laurent/Projects/2020_FairLearning/2020_W2_penalty_in_ML/2020_W2reg/Res

We then tested our method on a semi-synthetic image dataset, where we explicitly controlled the bias in the training data. To do so, we used 20000 training images and 8000 test images out of  the classic MNIST dataset\footnote{http://yann.lecun.com/exdb/mnist/}. Each image  $x_i$ has $28 \times 28$ pixels  and represents a handwritten digit. We considered in our experimental protocol that the images $x_i$ representing digits strictly lower than 5 (\textit{i.e.} 0, 1, 2, 3 or 4) have an output $y_i=0$ and the other images (\textit{i.e.} representing 5, 6, 7, 8 or 9) have an output  $y_i=1$. Note that the frequency of each digit was relatively stable in these datasets. Each digit was indeed represented between 1889 and 2232 times in the training set and between 757 and 905 times in the test set.
We then randomly draw in the training set and the test set a label $s_i \in \{0,1\}$ for each observation. A Bernoulli law with $p=0.5$ was used to draw $s_i$, so there is roughly the same amount of observations with  $s_i=0$ and $s_i=1$. All observations with $s_i=0$ were then rotated by 180 degrees.

In order to automatically predict the outputs $y_i$ using the images $x_i$, we used a basic Convolutional Neural Network (CNN) architecture with three stacks of Convolution/MaxPooling layers.
We then trained this classifier using 50 epochs and a batch size of 200 observations. The results were standard for this dataset with $99.8\%$ and $97.4\%$ good predictions on the training and test sets, respectively. On the test set, the true positive rate  (TP) was $0.963$ and the false positive rate (FP) was $0.0144$.

\subsubsection{Adding a bias in the training data}

We now impaired the training data in the group $S=0$ only in order to control how the trained classifiers will be biased. The test set remained unchanged. Two strategies were used to do so:
\textbf{Semi-Random (SR)} We randomly set $y_i=0$ to $65\%$ of the observations representing the digit 7 in the group $S=0$,
\textbf{Separate-Treatment (ST)} We first coarsely trained our classifier on the training set of Section~\ref{ssec:ExpProUnbiased} with 2 epochs and we then set $y_i=0$ to the $30\%$ of observations with $s_i=0$ and originally $y_i=1$ that had the lowest predictions. Remark that about $60\%$ of the transformed observations were true positives, so among the digits higher or equal to 5 in the group $S=0$ of the training set, the $18\%$ being considered as the most similar to 0, 1, 2, 3 or 4 were mislabelled.

In can be remarked that the two kind of biases simulated here corresponds to the \textit{Cause 1} of algorithmic bias in Section~\ref{ssec:AlgorithmicBiasMachineLearning}.
As our goal is to correct a disparate impact in two groups of images which are supposed to have the same distribution of output predictions, we use the regularization energy of Section~\ref{ssec:w2reg}. The regularization therefore favors similar predictions.

\subsubsection{Results}

\begin{table*}
\caption{Impact of the regularization on the generalization properties of two trained classifiers using the experimental protocol (SR) of Section~\ref{ssec:MNISTdataset}.}
\begin{center}
\begin{tabular}{|c||c|c|c||c|c|c||c|c|c|c|} \hline
                  & \multicolumn{3}{c||}{Train}     & \multicolumn{3}{c||}{Test}            &  \multicolumn{4}{c|}{Test}                 \\ \hline
              &  DI     &  GP0     &  GP1     &  DI      &  GP0         &  GP1      &   TP0   &    TN0    &   TP1   &    TN1    \\        \hline  \hline
$\lambda=0.$   &  0.87   & 1.00     & 1.00     &  0.86    &  0.92        &  0.96     &   0.83  &    0.97   &   0.96  &    0.97   \\        \hline    % SR  /  lambda = 0.
$\lambda=2.0e-5$ &  0.98   & 0.94     & 0.96     &  0.96    &  0.93        &  0.95     &   0.91  &    0.95   &   0.93  &    0.97   \\        \hline    % SR  /  lambda = 0.000020
\end{tabular}
\end{center}
\label{tab:ResMNIST2}
\end{table*}

The main results obtained using the experimental protocols \textit{SR} and \textit{ST} are presented in Fig.~\ref{fig:ResMNIST1}. In both cases, it is clear that increasing the influence of the regularization  term (\textit{i.e.} $\lambda$) leads to gradually increased DIs, until this influence is too high ($\lambda>2.0e-5$ in our tests). In this case, the trained decision rules start to significantly loose their predictive power, so they are useless.  We can also remark that the portion of good predictions in the groups $S=0$ and $S=1$ are particularly similar when the DI is close to $1$. This is particularly true when using the protocol protocol \textit{SR}.

Our nicest results here is that the gain of fairness we obtained by using regularized predictions came with an improved predictive power for  \textit{SR} and no loss of predictive power for \textit{ST}. The regularization then improved the generalization properties of the trained neural networks in these experiments.

We further explain this phenomenon by discussing the detailed results of Table~\ref{tab:ResMNIST2}. It represents the DI, and the portion of good predictions in the groups $S=0$ and $S=1$  (GP0 and GP1) obtained on the training and the test sets using two training strategies:  experimental protocol \textit{SR} with $\lambda=0$ (no regularization) and $\lambda=2.0e-5$ (good level of regularization according to Fig.~\ref{fig:ResMNIST1}). True positive (TP) and true negative (TN) rates are also given on the test set in the groups $S=0$ and $S=1$.

As expected, the trained neural network with no regularization is very accurate on the training set and its DI reflects the simulated bias in the training set. This DI is stable when generalizing on the test set but the predictions  are less accurate. When using our regularization strategy,  the DI becomes very close to 1 on the training set and the predictions  remain reasonably accurate. Contrary to the non-regularized cases, both the prediction accuracy and the DI are however stable when generalizing to the test set. This leads to a similar accuracy as when using no regularization but a clearly improved DI. Note that the gain of fairness is mainly due to an improved true positive rate in the group $S=0$.

\subsection{CelebA dataset}\label{ssec:CelebAdataset}

%RESULTS IN: /home/laurent/Projects/2020_CelebA

\subsubsection{Experimental protocol}\label{ssec:ExpProCelebA}

We now present more advanced results obtained on the \textit{Large-scale CelebFaces Attributes (CelebA)} Dataset\footnote{http://mmlab.ie.cuhk.edu.hk/projects/CelebA.html} \cite{liu2015faceattributes}. It contains more than 200K celebrity images, each with 40 binary attribute annotations. Contrary to what we studied in Section~\ref{ssec:MNISTdataset}, the images of the CelebA dataset cover large pose variations and background clutter. This makes this dataset far more complex to study than the MNIST dataset. The binary attributes are for instance \textit{Eyeglasses},  \textit{PaleSkin},  \textit{Smiling}, \textit{Young},  \textit{Male} and \textit{Attractive}.

Interestingly, it is relatively simple to train a classifier detecting some objective attributes, such as  \textit{Eyeglasses}, using a modern Neural Network (NN) architecture. For instance, we reached in the test set more than $98\%$ accuracy when detecting whether the represented celebrities weared eyeglasses.
Other attributes, such as \textit{Attractive}, are more complex to handle because they are subjective. As shown later in our results, it is however relatively simple to have more than $84\%$ of good predictions when predicting  whether a person in the test set is considered as attractive. In practice, this suggests that the persons who labeled the data were relatively coherent when choosing who was attractive or not. The main issue with the fact that \textit{Attractive} is subjective  however comes to the fact that it may be influenced by other undesirable attributes. In this section, we then use our regularization strategy to limit the impact of the  attribute \textit{Young} when predicting whether someone is considered as attractive or not, while preserving as much as possible a good prediction accuracy.

Remark that the algorithmic bias we will tackle in this section may be due to different causes identified in Section~\ref{ssec:AlgorithmicBiasMachineLearning}. It is obviously related to the \textit{Cause 1} because of the subjectivity of the labeling process. It is also likely to be due to the confounding variables of \textit{Cause 3}, at least in hidden latent spaces. For instance, wearing eyeglasses is less frequent for young persons than older ones, making potentially the young persons with  eyeglasses assimilated to older persons. In a similar vein, there are about two $3.5$ times more persons being considered as young than older ones, leading potentially to the \textit{Cause 2} of bias. By using this dataset, which is representative of what can be used in industrial applications, we have then high chances to learn biased decision rules.

We used the ResNet-18 convolutional Neural Network (NN) architecture \cite{He2016DeepRL} to predict the attribute \textit{Attractive} based on the CelebA images. The sensitive variable $S$ used for the regularization was also the attribute \textit{Young}. Note that the original NN architecture was unchanged when implementing our strategy, as our regularization method is fully encoded in the loss function. We used to two proposed regularization strategies to favor similar predictions (see Section~\ref{ssec:w2reg}) and similar prediction errors (see Section~\ref{sssec:sim_error_rates}) in the groups $S=1$ (young) and $S=0$ (not young). In both cases, different classifiers were trained by using $\lambda=0$ (no regularization) and  gradually increasing $\lambda$s with values ranging between $0$ and $1e-2$.

Additional results giving the convergence of the training algorithm for different values of $\lambda$ are also presented in Appendix~\ref{sec:ImpactLambda}.

\begin{figure*}[h]
  \centering
  \includegraphics[width=0.99\linewidth]{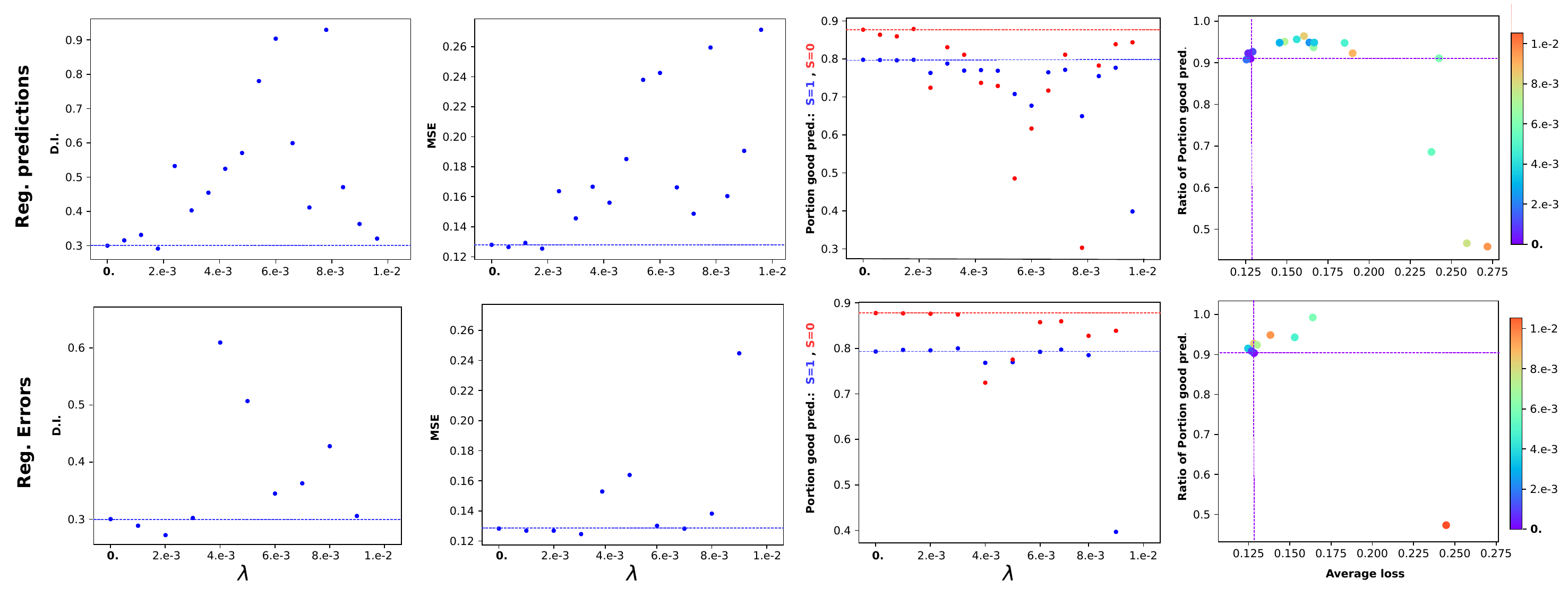}
  \caption{Results obtained on the CelebA dataset by favoring \textbf{(top)} similar predictions as in Section~\ref{ssec:w2reg} and \textbf{(bottom)} similar prediction errors as in Section~\ref{sssec:sim_error_rates}. In both cases, different classifiers where trained with $\lambda=0$ (no regularization) and   strictly positive $\lambda$s between $0$ and $1e-2$. We represent, the disparate impact (DI), Mean Squared Error (MSE), the portion of good predictions in the groups $S=0$ and $S=1$  (GP0 and GP1) obtained  on the test set  using each trained classifier, as well as the average ratios GP0/GP1. The dashed lines in each subfigure show the average scores obtained with no regularization.  }
  \label{fig:ResCelebA1}
\end{figure*}

\subsubsection{Results}\label{ssec:ResultsCelebA}

The main results obtained on the CelebA dataset are presented in Fig.~\ref{fig:ResCelebA1}.
As in Fig.~\ref{fig:ResMNIST1}, it represents the disparate impacts (DI) the mean squared errors (MSE) and the portion of good predictions for the images in the groups $S=0$ and $S=1$ (GP0 and GP1) obtained on the test set. To efficiently compare the gain of fairness with the loss of prediction accuracy, we also present a point cloud representing the MSE and the minimal value between $GP0/GP1$ and $GP1/GP0$ of each trained classifier.

The global behavior of our strategy is similar to what we observed in Section~\ref{ssec:MNISTdataset} on the MNIST dataset.
When increasing $\lambda$ until about $5.5e-3$, the results are more and more fair, \textit{i.e.} the DI is higher and higher, and GP1 is closer and closer to GPO. It appears that a DI of 0.6  can be reached with little loss of predictive power in this test,  and  that $GP0$ can be very close to $GP1$ under this constraint.
It can also be remarked that for $\lambda<5.5e-3$, favoring similar predictions (see Section~\ref{ssec:w2reg}) or similar prediction errors (see Section~\ref{sssec:sim_error_rates}) in the groups $S=0$ and $S=1$ had a relatively similar effect. When using the method of Section~\ref{ssec:w2reg} with $\lambda>5e-3$, the predictions are however unstable: They are sometimes similar to those obtained using $\lambda \approx 4.5e-3$, sometimes very fair ($DI>0.8$) but inaccurate ($MSE>0.2$), and sometimes clearly unfair ($DI  \approx 0.3$) and inaccurate ($MSE>0.2$).
When using the method of Section~\ref{sssec:sim_error_rates} with $\lambda>6e-3$, the predictions first appear as similar to those obtained without regularization. Then, they  become more unstable for  $\lambda>8.5e-3$, as for the results represented with $\lambda>8.5e-3$.
Another point to emphasize is that although a DI of about 0.6 can be reached with little loss of predictive power, DIs above 0.8 can only be reached with strong loss of predictive power. This is interestingly not the case for the difference between the portion of good predictions GP0 and GP1 (which takes into account the true prediction $\hat{Y}$). The ratio $GP0/GP1$ is indeed very close to 1 when favoring similar prediction errors with $\lambda>5e-3$.
Importantly, the regularization finally allowed to get clearly more fair results with little loss of predictive power for weights $\lambda \in [3.5e-3 , 5.5e-3]$.

\begin{figure*}[h]
  \centering
  \includegraphics[width=0.99\linewidth]{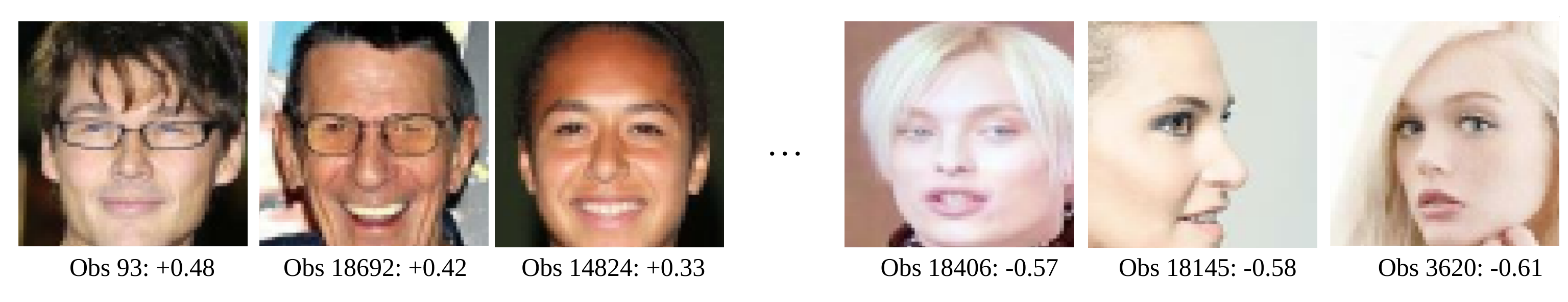}
  \caption{Observations with the highest differences of predictions with or without regularization. We recall that $Y$ represents the subjective variable \textit{Attractive},
  which represents who is attractive for the persons who labelled the data, and $S$ is the variable \textit{Young}.
  }
  \label{fig:ResCelebA2}
\end{figure*}

It is now interesting to  discuss what was the practical impact of the regularization algorithm in these tests. We recall that $Y$ represents the subjective attribute \textit{Attractive}, which represents who is attractive for the persons who labelled the data and that $S$ is the attribute \textit{Young}.
We then measured the scores differences  $f_{\bar{\theta}}(x_i)-f_{\tilde{\theta}}(x_i)$, where the parameters $\bar{\theta}$ were obtained by favoring similar error rates with
$\lambda = 4.e-3$ and the parameters $\tilde{\theta}$ were obtained without regularization.
The three highest positive and negative differences between $f_{\bar{\theta}}(x_i)$ and $f_{\tilde{\theta}}(x_i)$ are represented in Fig.~\ref{fig:ResCelebA2}
This figure makes clear that the regularization with $S$ encoding the attribute \textit{Young} made more attractive the persons with eyeglasses and relatively dark tans, and made less attractive the women with very pale skins and no eyeglasses. In addition to the fact that the \textit{young} images are positively correlated with \textit{attractive} and  \textit{female} in the training set, this can be explained by different reasons: The frequency of persons with eyeglasses is 4.71 times more frequent in \textit{old} persons than \textit{young} ones; \textit{Males} are 2.01 time more frequent in \textit{old} persons than \textit{young} ones; \textit{Pale skins} are 1.8 times more frequent in \textit{young} persons than \textit{old} ones. In all cases, this suggests that the regularization has lead to coherent results with regard to the distribution of the attributes in the training set.

\section{Discussion}

In this paper, we have proposed a new method to temper the algorithmic bias in Neural-Network based classifiers. The first key advantage of this strategy is that it can be integrated to any kind of Neural-Network architecture, as it only overloads the loss term when training optimal decision rules. This additionally makes it straightforward to integrate it to existing deep learning solution. As demonstrated on the CelebA dataset, its second key advantage is that it scales particularly well to large image datasets, which are increasingly ubiquitous in industrial applications of artificial intelligence.
In terms of methodology, the central idea of this work was to define two alternatives of a  fairness penalty term which can be naturally used in stochastic gradient descent based optimization algorithms. The first alternative favors similar prediction outputs in two groups of data and the second one favors similar prediction errors in these groups. The main technical lock we had to address was to approximate the gradient of this penalty term in the real-life context where the size of mini-batches can be relatively small.
Our results have shown the good properties our regularization strategy.

Future work will first consist in extending our strategy to the non-binary classification case, which should be straightforward by using one-hot-encoding representations of the outputs. A more ambitious extension, with potentially numerous applications, would be to address the regression case too.  We may also explore the use of this strategy on different kinds of data, such as those treated in Natural Language Processing, as it can be used on any kind of data. In order to make the optimization process more stable when the algorithmic bias is particularly complex, a promising strategy would be finally to work on the latent spaces of the neural-network hidden layers and not the outputs directly.

%The github repository with our implementation of the proposed strategy will be finally made public after paper acceptance.

\bibliographystyle{unsrt}

\appendix

\section{Proof of G\^ateaux Differentiability}\label{sec:ProofPropGradW2obs}

\subsection{Proof of Proposition~\ref{prop:GateauxDiffTransportCost}}

\begin{proof}
Recall from the duality of the transport problem that it can be rewritten as the optimization problem
\begin{equation}\label{dual}
    \mathcal{W}^2_2(\mu,\nu)=\sup_{f}\int x^2-2f(x)d\mu(x)+\int y^2-2f^*(y)d\nu(y),
\end{equation}
where the optimization is over the set of convex functions. Following \cite{RockafellarBook1970}, we then denote here $f$ a convex function and $f^*$ its conjugate. Let $f_0$ and $f_{t}$ be the solutions for $\mathcal{W}^2_2(\mu,\nu)$ and $\mathcal{W}^2_2(\mu+t\alpha,\nu+t\beta)$. We then have
\begin{equation*}
    \mathcal{W}^2_2(\mu+t\alpha,\nu+t\beta)-\mathcal{W}^2_2(\mu,\nu)
    \leq  t\left( \int x^2-2f_t(x)d\alpha(x)+\int y^2-2f_t^*(y)d\beta(y)\right)
\end{equation*}
and
\begin{equation*}
    \mathcal{W}^2_2(\mu+t\alpha,\nu+t\beta)-\mathcal{W}^2_2(\mu,\nu)
    \geq t\left( \int x^2-2f_0(x)d\alpha(x)+\int y^2-2f_0^*(y)d\beta(y)\right).
\end{equation*}
Note that, following \cite{del2019central,LoubesII19}, the convergence $\mu+t\alpha\xrightarrow{w} \mu$ implies that $f_t\rightarrow f_0$ uniformly on the compact sets of the support of $\mu$ if the support is connected. As a consequence, if $\operatorname{Supp}(\alpha)\subset \operatorname{Supp}(\mu)$ and  $\operatorname{Supp}(\beta)\subset \operatorname{Supp}(\nu)$, then under the assumption of compact supports we have

\begin{equation}\label{gateaux2}
   D\mathcal{W}^2_2(\mu,\nu)(\alpha,\beta)= \int x^2-2f_0(x)d\alpha(x)+\int y^2-2f_0^*(y)d\beta(y).
\end{equation}
Recall from \cite{VillaniBook2003} (Section 2.2.) that in the real line case the potential of the transport, solution of Eq.~\eqref{dual}, is a primitive of the transport map, which is the map $F^{-1}_{\mu}(F_{\nu})$. Then the conclusion becomes straightforward.
\end{proof}

\subsection{Proof of Lemma~\ref{gat:los}}

\begin{proof}
Set $t>0$ and $\beta$, then compute
\begin{align*}
    L(\mu+t\beta)-L(\mu)=  \int l d \mu+ t\int l d \beta- \int l d \mu=t\int l d \beta.
\end{align*}
In consequence we have that
\begin{align*}
    \frac{L(\mu+t\beta)-L(\mu)}{t}=  \int l d \beta.
\end{align*}
\end{proof}

\section{Extension to error rates Eq.~\eqref{eq:finalW2gradApprox_err}}\label{sec:ProofgradW2squaredErrors}

By using the same strategy as in
Proposition~\ref{prop:GateauxDiffTransportCost}
with $(f_{\theta}(x_i)-y_i)^2$ and the $\tilde{H}_g$ instead of $f_{\theta}(x_i)$ and the $H_g$, we can compute:
\begin{align}
& \displaystyle{
\Delta_{\tau} \left[
  \mathds{1}_{s_i=0}
 \frac{     (f_{\theta}(x_i)-y_i)^2 - cor_1((f_{\theta}(x_i)-y_i)^2)   }{ n_0 \left(\tilde{H}_{0}^{j_i+1} - \tilde{H}_{0}^{j_i} \right)}
   \right.} \nonumber \\
  &
   \displaystyle{
     \left. -   \mathds{1}_{s_i=1}
      \frac{     cor_0((f_{\theta}(x_i)-y_i)^2) - (f_{\theta}(x_i)-y_i)^2 }{ n_1 \left(\tilde{H}_{1}^{j_i+1} - \tilde{H}_{1}^{j_i} \right)}
     \right] \,.
   }
\end{align}.

This equation specifically expresses how to back-propagate  the impact of a squared error  $(f_{\theta} (x_i)-y_i)^2$, and not an output observation $f_{\theta} (x_i)$, on
$W_2^2 (\tilde{\mu}^n_{\theta,0},\tilde{\mu}^n_{\theta,1})$. From Lemma~\ref{gat:los}, we know that the chain rule applies on the G\^ateaux differentiability which is used to back-propagate the impact of $f_{\theta} (x_i)$ on the distributions.
As
\begin{equation}
\frac{\partial  (f_{\theta}(x_i)-y_i)^2}{\partial  f_{\theta}(x_i)} = 2 (f_{\theta}(x_i)-y_i) \,,
\end{equation}
we can then simply deduce that the impact of an observation  $f_{\theta} (x_i)$ on
$W_2^2 (\tilde{\mu}^n_{\theta,0},\tilde{\mu}^n_{\theta,1})$ can be back-propagated using Eq.~\eqref{eq:finalW2gradApprox_err}:

\begin{align}
&\displaystyle{
2 \Delta_{\tau}
\left[
 \mathds{1}_{s_i=0}
 \frac{    (f_{\theta}(x_i)-y_i)^2 - cor_1 \left((f_{\theta}(x_i)-y_i)^2\right)   }{n_0 \left(\tilde{H}_{0}^{j_i+1} - \tilde{H}_{0}^{j_i}\right) \left(f_{\theta}(x_i)-y_i\right)^{-1}}
 \right. } \nonumber  \\
&\displaystyle{
\left.
  - \mathds{1}_{s_i=1}
\frac{    cor_0 \left((f_{\theta}(x_i)-y_i)^2 \right) - (f_{\theta}(x_i)-y_i)^2 }{n_1 \left( \tilde{H}_{1}^{j_i+1} - \tilde{H}_{1}^{j_i}\right) \left(f_{\theta}(x_i)-y_i\right)^{-1} }
\right] \,.
}
\end{align}

\section{Extensions of the method of Section~\ref{ssec:QuickW2inBatch}}\label{sec:ExtensionsQuickW2inBatch}

\subsection{Wasserstein-1 distances}\label{sssec:W1_alternative}

Our approach can be straightforwardly extended to approximate Wasserstein-1 distances.
By using the same reasoning as in Section~\ref{ssec:QuickW2inBatch}, it can be shown
that the impact of an observation  $f_{\theta} (x_i)$ on
$W_1 (\tilde{\mu}^n_{\theta,0},\tilde{\mu}^n_{\theta,1})$ can be back-propagated using
\begin{equation}\label{eq:finalW1gradApprox}
\Delta_{\tau}  \displaystyle{ \left[
 \mathds{1}_{s_i=0}
 \frac{  sign \left(   \eta^{j_{i}} - \eta^{j_{i}'} \right)}{n_0 (H_{0}^{j_i+1} - H_{0}^{j_i})}   - \mathds{1}_{s_i=1}
\frac{  sign \left(   \eta^{j_{i}'} - \eta^{j_{i}} \right)}{n_1 (H_{1}^{j_i+1} - H_{1}^{j_i})}
\right]} \,,
\end{equation}
instead of Eq.~\eqref{eq:finalW2gradApprox},
where $sign(x)$ is equal to $+1$ or $-1$ depending on the sign of $x$.
We emphasize that the distances between the cumulative densities are therefore not taken into account when computing the gradients of the Wasserstein-1, although this is the case for  Wasserstein-2 distances.

\subsection{Logistic Regression}\label{sssec:LR_alternative}

We now show how to simply implement our regularization model for Logistic Regression. We minimize:
\begin{equation}\label{eq:LogitReg}
\hat{\theta} = \argmin_{\theta}
     \frac{1}{n} \sum_{i=1}^n \log \left( f_{\theta} (x_i)^{y_i} \left( 1 - f_{\theta} (x_i) \right)^{1-y_i} \right) \nonumber 
   + \lambda W_2^2 (\mu^n_{\theta,0},\mu^n_{\theta,1})  \,,
\end{equation}
where $f_{\theta} (x_i) = (1+\exp{( -\theta^0 -\theta' x_i)})^{-1}$ is  the logistic function and $\theta = (\theta^1, \ldots , \theta^p)'$ is a vector in $\mathbb{R}^p$ representing the weights given to each dimension. The derivatives of the whole energy Eq.~\eqref{eq:LogitReg} with respect to each $\theta^j$, $j=0, \ldots , p$, can be directly  computed using finite differences here.

We emphasize that a fundamental difference between using our Wasserstein based regularization model in Section~\ref{ssec:QuickW2inBatch} and here is that $p$ derivatives of the minimized energy are approximated using Logistic Regression (derivatives w.r.t. the $\theta^j$, $j=0, \ldots , p$), while  $n$ derivatives  are required when using Neural Networks with a standard gradient descent (derivatives w.r.t. the $f_{\theta} (x_i)$, $i=0, \ldots , n$). As a cumulative histogram is computed each time the derivative of a Wasserstein-2 distance is approximated, this task can be bottleneck for common Neural-Networks applications where $n$ is large.
This fully justifies the proposed batch-training regularization strategy of Section~\ref{ssec:QuickW2inBatch}.

\subsection{Automatic tuning of $\lambda$}\label{sssec:autoTuneLambda}

The minimized energy Eq.~\eqref{eq:minimizedEnergy}  depends on a weight $\lambda$ which balances the influence of the regularization term $W_2^2 (\mu^n_{\theta,0},\mu^n_{\theta,1}) $ with respect to the data attachment  term $R(\theta)$.
A simple way to automatically tune $\lambda$ is the following. Compute the average derivatives of $W_2$ and $R$ after a predefined warming phase of several epochs, where $\lambda=0$. We denote $d_{W_2}$ and $d_{R}$ these values. Then tune $\lambda$ as equal to $\alpha \frac{g_{R}}{g_{W_2}}$, where $\alpha$ is typically in $[0.1,1]$. This makes it intuitive to tune the scale of $\lambda$.

In the disparate impact case (Section~\ref{ssec:QuickW2inBatch}), it can be interesting to accurately adapt $\alpha$ to the machine learning problem, in order to  finely tune $\lambda$ with regards to the fact that we simultaneously want fair and accurate  predictions. Inspired by the hard constraints of \cite{ZafarICWWW17} to enforce fair predictions, we update $\alpha$ based on measures of the Disparate Impact (DI),
Eq.~\eqref{def:DIclassifier}, and average Prediction Accuracy (Acc) at the beginning of each epoch. Remark that lowering the Wasserstein-2 distance between the predictions $f_{\theta}(x_i)$ in groups $0$ and $1$ naturally tend to make decrease $\mathbbm{1}_{f_{\theta}(x_i,s_i=0)>0.5}-\mathbbm{1}_{f_{\theta}(x_i,s_i=1)>0.5}$, which we empirically verified. The disparate impact therefore tends to be improved. We believe that hard constraints based on other fairness measures could also be used. Establishing a clear relation of causality between the Wasserstein-2 distance and different fairness measures is however out of the scope of this paper and hence considered as future work.
Note that the same technique holds in the squared error case (Section~\ref{sssec:sim_error_rates}) but
the Disparate Mean Squared Error (DMSE) index, Eq.~\eqref{def:DEclassifier}, is used instead of the DI.

In the experiments of Section~\ref{ssec:Adultdataset}, our hard constraints are for instance:
If the prediction accuracy is too low (Acc$<0.75$), then $\alpha$ is slightly decreased to favor the predictions accuracy ($\alpha=0.9\alpha$). If the prediction accuracy is sufficiently high and the DI is too low (DI$<0.85$) then $\alpha$ is slightly increased ($\alpha=1.1\alpha$) to favor fair decisions.
We empirically verified in our experiments that $\alpha$ converges to satisfactory values using this method, if the classifier is able to learn classification rules leading to sufficiently high PA. Parameter $\alpha$ converges to zero otherwise.

\section{Impact of $\lambda$ on the convergence}\label{sec:ImpactLambda}

\subsection{Results}

\begin{figure*}[h]
  \centering
  \includegraphics[width=0.99\linewidth]{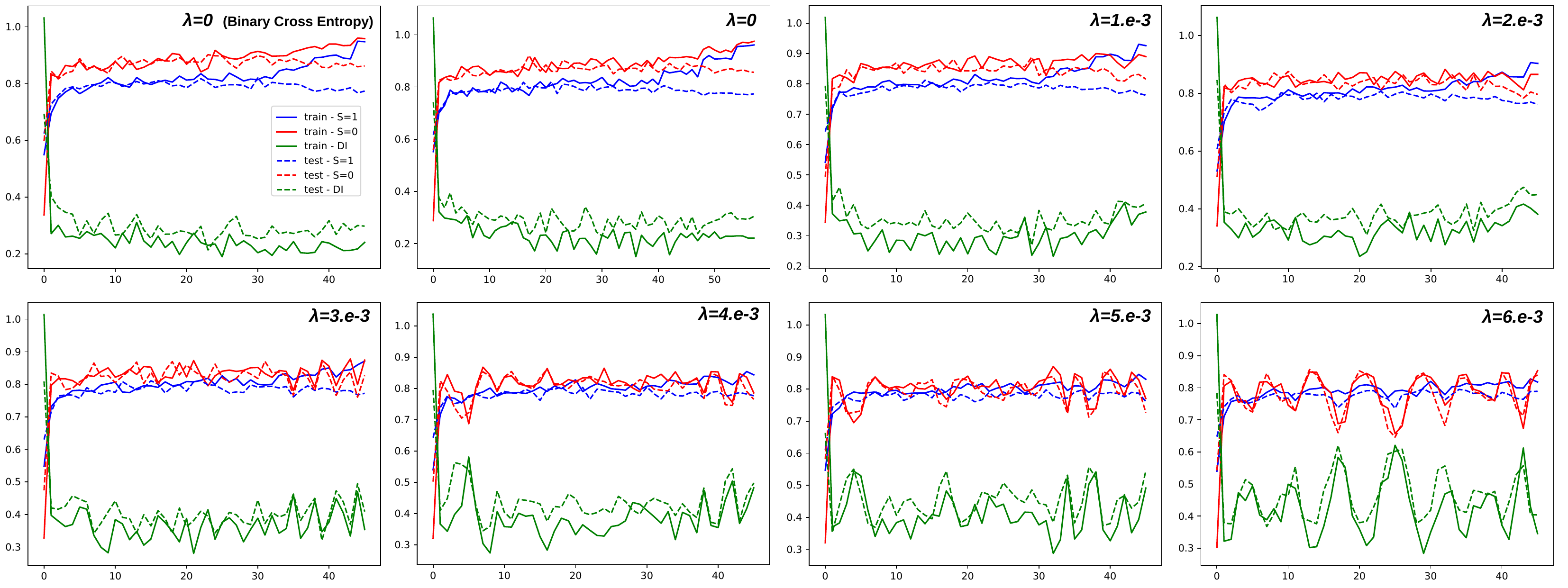}
  \caption{
  Impact of $\lambda$ on the convergence of the neural-network predictions on the CelebA dataset, when using the experimental protocol of Section \label{ssec:ExpProCelebA} . All results were obtained using the mean squared error as a loss, except for the sub-figure on the top-left, which were obtained using Binary Cross Entropy.
  Blue and red curves represent the accuracy for the observations with $S=1$ and $S=0$, respectively. Green curves represent the disparate impact. Continuous and dashed curves were finally obtained on the training and test sets. The iterations represented in abscissa are in mini-batches and not in epochs.}
  \label{fig:CelebA_convergences}
\end{figure*}

In this section, we extend the results of Section~\ref{ssec:ExpProCelebA} by discussing the convergence of the training algorithm on the training set and the test set for different values of $\lambda$.
Each sub-figure specifically represents the accuracy of the prediction model mini-batch after mini-batch, where we distinguish the average accuracy obtained on  observations with $S=0$ and those with $S=1$. The disparate impact is also given.
We additionally compare the convergence of the training algorithm without regularization when using a mean squared error (MSE) loss, as in other experiments, and a binary cross entropy (BCE) loss, which is known to be less sensitive than MSE to outlier observations. The regularization was applied on the output predictions directly.  Results are detailed in Fig.~\ref{fig:CelebA_convergences}.

\subsection{Discussion}

We can first notice that the convergence properties of the training algorithm with $\lambda=0$ are very similar when using the MSE and the BCE loss. The potential outlier observations seem therefore to have no or little impact here. In both cases and both for $S=0$ and $S=1$, the observed level of accuracy is similar for the training set and the test set until about the 30th iteration. Then, the accuracy becomes clearly higher on the training set than on the test set, which means that the trained model overfits the training data. This phenomenon is particularly strong for the observations in the group $S=1$, which have a lower accuracy than in the group $S=0$ before overfitting.  In all cases, the disparate impact is close to $0.25$ at convergence.

Now, when increasing $\lambda$ from $1.e-3$ to  $5.e-3$ the level of accuracy becomes increasingly  similar between $S=0$ and $S=1$, and the disparate impact is closer and closer to $0.4$. It can be remarked that for $\lambda=4.e-3$ and   $\lambda=5.e-3$, the curves start oscillating when the training algorithm overfits the training data. The regularization therefore seems to be strong compared with the data attachment term. This phenomenon is stronger, even before overfitting, for $\lambda=6.e-3$. This explains why the results become unstable in Fig.~\ref{fig:ResCelebA1}-(top) for high values of $\lambda$, and confirms that a reasonable trade-off should be searched between regularization and prediction accuracy when tuning $\lambda$.

\end{document}